\documentclass[letterpaper]{article} % DO NOT CHANGE THIS
\usepackage{aaai24}  % DO NOT CHANGE THIS
\usepackage{times}  % DO NOT CHANGE THIS
\usepackage{helvet}  % DO NOT CHANGE THIS
\usepackage{courier}  % DO NOT CHANGE THIS
\usepackage[hyphens]{url}  % DO NOT CHANGE THIS
\usepackage{graphicx} % DO NOT CHANGE THIS
\usepackage{natbib}  % DO NOT CHANGE THIS AND DO NOT ADD ANY OPTIONS TO IT
\usepackage{caption} % DO NOT CHANGE THIS AND DO NOT ADD ANY OPTIONS TO IT
\usepackage{algorithm}
\usepackage{algorithmic}
\usepackage{subfigure}
\usepackage{multirow}
\usepackage{makecell}
\usepackage{amsthm}
\usepackage{amsmath}
\usepackage{amsfonts,amssymb}
\usepackage{bm}
\usepackage{booktabs}

\usepackage{newfloat}
\usepackage{listings}
\DeclareCaptionStyle{ruled}{labelfont=normalfont,labelsep=colon,strut=off} % DO NOT CHANGE THIS
\floatstyle{ruled}
\newfloat{listing}{tb}{lst}{}
\floatname{listing}{Listing}
\title{Homophily-Related: Adaptive Hybrid Graph Filter for Multi-View Graph Clustering}
\author{
    Zichen Wen\textsuperscript{\rm 1},
    Yawen Ling\textsuperscript{\rm 1},
    Yazhou Ren\textsuperscript{\rm 1,\rm 2}\thanks{Corresponding author.},
    Tianyi Wu\textsuperscript{\rm 1},
    Jianpeng Chen\textsuperscript{\rm 3},
    Xiaorong Pu\textsuperscript{\rm 1,\rm 2},\\
    Zhifeng Hao\textsuperscript{\rm 4},
    Lifang He\textsuperscript{\rm 5}
}
\affiliations{
    \textsuperscript{\rm 1}School of Computer Science and Engineering, University of Electronic Science and Technology of China, Chengdu, China\\
    \textsuperscript{\rm 2}Shenzhen Institute for Advanced Study, University of Electronic Science and Technology of China, Shenzhen, China\\
    \textsuperscript{\rm 3}Department of Computer Science, Virginia Tech, Blacksburg, VA, USA\\
    \textsuperscript{\rm 4}College of Science, Shantou University, Shantou, China\\
    \textsuperscript{\rm 5}Department Computer Science and Engineering, Lehigh University, Bethlehem, PA, USA\\
    \{Zichen.Wen, yawen.Ling, TianYi-Wu\}@outlook.com, 
    \{yazhou.ren, puxiaor\}@uestc.edu.cn, \\
    jianpengc@vt.edu, 
    haozhifeng@stu.edu.cn, lih319@lehigh.edu
}

\usepackage{bibentry}
\begin{document}

\maketitle

\begin{abstract}
Recently there is a growing focus on graph data, and multi-view graph clustering has become a popular area of research interest. Most of the existing methods are only applicable to homophilous graphs, yet the extensive real-world graph data can hardly fulfill the homophily assumption, where the connected nodes tend to belong to the same class. Several studies have pointed out that the poor performance on heterophilous graphs is actually due to the fact that conventional graph neural networks (GNNs), which are essentially low-pass filters, discard information other than the low-frequency information on the graph. Nevertheless, on certain graphs, particularly heterophilous ones, neglecting high-frequency information and focusing solely on low-frequency information impedes the learning of node representations. To break this limitation, our motivation is to perform graph filtering that is closely related to the homophily degree of the given graph, with the aim of fully leveraging both low-frequency and high-frequency signals to learn distinguishable node embedding. In this work, we propose \textbf{A}daptive \textbf{H}ybrid \textbf{G}raph \textbf{F}ilter for Multi-View Graph \textbf{C}lustering (AHGFC). Specifically, a graph joint process and graph joint aggregation matrix are first designed by using the intrinsic node features and adjacency relationship, which makes the low and high-frequency signals on the graph more distinguishable. Then we design an adaptive hybrid graph filter that is related to the homophily degree, which learns the node embedding based on the graph joint aggregation matrix. After that, the node embedding of each view is weighted and fused into a consensus embedding for the downstream task. Experimental results show that our proposed model performs well on six datasets containing homophilous and heterophilous graphs.
\end{abstract}

\section{Introduction}
Multi-view graph clustering (MVGC) has become a popular research topic due to the considerable structural information contained in graph data. Some related research works have been proposed and achieved impressive results, such as O2MAC~\cite{o2multi}, graph convolutional encoders~\cite{XiaWYGHG22}, and variational graph generator for multi-view graph clustering~\cite{chen2022variational}. 
% \begin{figure}[h]
%     \centering
%     \subfigure{
%         \includegraphics[height=1.2in,width=1.60in]{A_ppt.pdf}}
%     %\hspace{0.5in}
%     \subfigure{
%         \includegraphics[height=1.2in,width=1.60in]{S_ppt.pdf}}
%     \caption{Eigenvalues of adjacency matrix $\mathbf{A}$ (left) and the proposed graph joint aggregation matrix $\mathbf{S}$ (right) on the second graph of the ACM dataset. The boundary between low and high frequencies is unclear in $\mathbf{A}$, while $\mathbf{S}$ enhances the distinguishability between them.}
%     \label{fig:eigenvalues}
% \end{figure}
% \begin{figure}[!t]
%     \centering
%     \subfigure{
%         \includegraphics[height=1.36in,width=3.3in]{CameraReady/LaTeX/A_S_lyw.pdf}}
%     % \subfigure{
%     %     \includegraphics[height=1.35in,width=1.59in]{CameraReady/LaTeX/A.pdf}}
%     % \subfigure{
%     %     \includegraphics[height=1.35in,width=1.62in]{CameraReady/LaTeX/S.pdf}}
%     \caption{Eigenvalues of adjacency matrix $\mathbf{A}$ (left) and the proposed graph joint aggregation matrix $\mathbf{S}$ (right) on the second graph of the ACM dataset. The boundary between low and high frequencies is unclear in $\mathbf{A}$, while $\mathbf{S}$ enhances the distinguishability between them.}
%     \label{fig:eigenvalues}
% \end{figure}
Undoubtedly, these approaches have contributed to the development of the MVGC. These approaches, however, are based on the homophily assumption, \textit{i.e.}, the nodes connected by edges belong to the same class~\cite{Hamilton2020GraphRL, wang2022powerful}. Yet, not all graphs in the real world exhibit homophily. 
% Due to the fact that graph data in real scenarios are often difficult to satisfy this condition, more often than not, edges are preferred to connect nodes of different classes, \textit{i.e.}, heterophilous graphs. 
For example, chemical interactions in proteins often occur between different types of amino acids~\cite{zhu2020beyond}. Nt et al. and~\citeauthor{li2019label} point out that the GNNs based on the homophily assumption are actually low-pass filters spectrally~\cite{nt2019revisiting, li2019label}.
It is well known that low-pass filters employ truncation at a certain frequency to obtain low-frequency signals (smooth signals), \textit{i.e.}, similarity information, on the graph.
However, the operational principle of the low-pass filters neglects the high-frequency signals (non-smooth signals) on the graph, that is, dissimilarity information.
This is feasible on homophilous graphs primarily composed of low-frequency signals, however, on heterophilous graphs dominated by high-frequency signals, capturing solely low-frequency signals is insufficient to facilitate the learning of node representations. 
\citeauthor{bo2021beyond} have also pointed out that capturing high-frequency signals on heterophilous graphs is more valuable for learning great node
representations~\cite{bo2021beyond}.

\begin{figure}[!t]
    \centering
    \subfigure{
        \includegraphics[height=1.35in,width=3.3in]{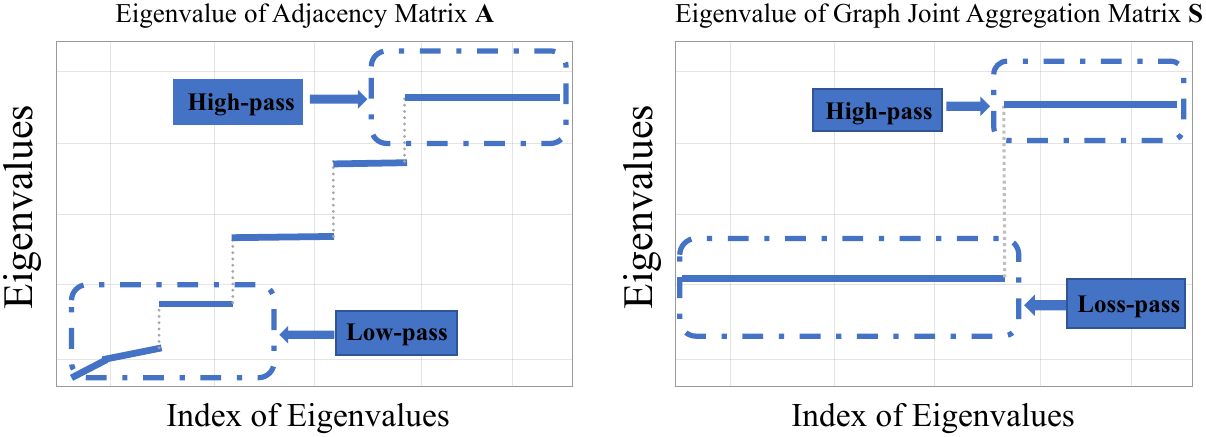}}
    % \subfigure{
    %     \includegraphics[height=1.35in,width=1.59in]{CameraReady/LaTeX/A.pdf}}
    % \subfigure{
    %     \includegraphics[height=1.35in,width=1.62in]{CameraReady/LaTeX/S.pdf}}
    \caption{Eigenvalues of adjacency matrix $\mathbf{A}$ (left) and the proposed graph joint aggregation matrix $\mathbf{S}$ (right) on the second graph of the ACM dataset. The boundary between low and high frequencies is unclear in $\mathbf{A}$, while $\mathbf{S}$ enhances the distinguishability between them.}
    \label{fig:eigenvalues}
\end{figure}
In general, the above analysis elucidates the reasons for the hindered performance of MVGC approaches based on GNNs acting as low-pass filters on heterophilous graphs from graph-filtering perspective: the loss of high-frequency information on the graph leads to inadequate information extraction. 
In order to solve the problem of information loss, some researchers propose to obtain abundant information on the graph by designing multiple graph filters~\cite{luan2022revisiting}.
Specifically, several works have proposed using hyperparameters to weigh and combine low-pass and high-pass filters, which are called mixed filter~\cite{Kangzhao}, as a way to obtain low and high-frequency information on the graph. However, since the graph type in MVGC is unknown, the hyperparameters as filter weights cannot be accurately aligned with the ratio of low and high-frequency signals on the graph, which again leads to loss of information on the graph.
Acknowledging the limitation of the hyperparameter weighting strategy, \citeauthor{bo2021beyond} further address the issue by incorporating an attention mechanism to facilitate an adaptive process for the hybrid graph filter~\cite{bo2021beyond}. Nevertheless, its adaptability primarily hinges on utilizing node feature information to explore the proportions between low-frequency and high-frequency signals and generates a large number of parameter calculations.

After thoroughly examining existing MVGC methods and enhanced graph filtering techniques for heterophilous graphs, we observe that many approaches suffer from information loss due to their disregard of high-frequency signals. Even the hybrid filtering enhancements still exhibit information loss due to the absence of a suitable adaptive process.
To address this issue, we propose an adaptive hybrid graph filter with an adaptive mechanism related to the homophily degree. The homophily degree, $\textit{i.e.}$, the measure of similarity (smoothness) on the graph, directly reflects the proportion of low and high-frequency information on the graph and is computationally simpler compared to the adaptation using the attention mechanism. But it is difficult to measure the homophily degree of the given graph because of the absence of true label information. Alternatively, we try to utilize the overall embedding from the current iteration to measure the graphs from each view, \textit{i.e.}, homophily ratios ($hr$). Specifically, the adaptive mechanism of $hr$ control is as follows: when facing the homophilous graphs ($hr \to 1$), the adaptive hybrid graph filter mainly intercepts the low-frequency signals to capture the similarity information; when facing the low-homophilous graphs ($hr \to 0$), the adaptive hybrid graph filter mainly intercepts the high-frequency signals to capture the dissimilarity information. 

Moreover, we notice that nearly all graph filtering methods use the adjacency matrix $\mathbf{A}$ for filter construction. However, this may result in information loss or noise introduction because $\mathbf{A}$ is inherently unreliable, $\textit{e.g.}$, the signal frequency distribution is spread out or the boundaries between low and high-frequency signals are not clear. The frequency distribution of $\mathbf{A}$ is scattered, as shown on the left side of Fig.~\ref{fig:eigenvalues}. The filter that is constructed from $\mathbf{A}$ can only access a portion of the frequency bands boxed in the figure, resulting in a significant loss of information. To cope with this situation, some works attempt to design filters for each frequency band to reduce information loss~\cite{wu2022beyond}, but this increases model complexity due to training multiple filters. For this reason, rather than directly utilizing the adjacency matrix $\mathbf{A}$ to construct the adaptive hybrid graph filter, we first undertake a graph joint process using the encoded feature matrix $\mathbf{Z}_x$ and the adjacency matrix $\mathbf{Z}_a$ to extract the consensus information of the node features concerning the neighbor relationships. Following this, drawing inspiration from the aggregation operation, we employ the output of the graph joint process to create a graph joint aggregation matrix $\mathbf{S}$, which facilitates the subsequent filtering operation. As shown on the right side of Fig.~\ref{fig:eigenvalues}, the frequency distribution of $\mathbf{S}$ is concentrated and the low and high-frequency signals are more distinguishable.

In summary, our key contributions are as follows:
\begin{itemize}
    \item In this work, we investigate the challenges of MVGC for heterophilous graphs based on graph filtering. Our goal is to effectively and intelligently leverage both low and high-frequency information to prevent information loss and facilitate the learning of distinct nodes.
    \item To achieve our goal, we propose an adaptive hybrid graph filter. We construct this filter using the graph joint aggregation matrix that enhances the distinguishability between low and high frequencies. We design an adaptive mechanism related to the homophily degree, which can adaptively mine both low and high-frequency information to learn node representations with less complexity.
    \item Experiments on both homophilous and heterophilous graphs show that the proposed AHGFC effectively mitigates the performance limitations of MVGC on heterophilous graphs, and the evaluation metrics on some datasets even exceed those of known SOTAs.
\end{itemize}

\section{Related Works}
\subsection{Multi-View Graph Clustering}\label{sec: Related MVGC}
The goal of multi-view graph clustering is to find the partition of the graph data with multiple data providing adjacency relationship and nodes feature information.
As one of the MVGC methods, 
O2MAC is proposed to capture the shared feature representation by designing a One2Multi graph autoencoder.~\citeauthor{2020Contrastive} introduce a self-supervised model to learn node representations by contrasting structural views~\cite{2020Contrastive}. Inspired by contrastive learning, Multi-View Contrastive Graph Clustering (MCGC) gets smooth node representations by removing the high-frequency noise by filter, then learns a consensus graph regularized by graph contrastive loss~\cite{pan2021multi}. ~\citeauthor{XiaWYGHG22} propose a multi-view graph clustering network using Euler transformations to enhance node attributes and guide learning with clustering labels. However, these methods are based on the homophily assumption and ignore the importance of high-frequency information for node representations. This leads to their inability to perform well in the face of heterophilous graphs.

% 2. GNNs with spectral domain / MVHGC and spectral methods...\\
\subsection{Heterophilous Graph Learning}
The poor performance of GNNs on the heterophilous graphs is attributed to its abandonment of high-frequency signals, resulting in the inability to effectively aggregate node information. \citeauthor{2021Beyond} analyze the roles of high-frequency and low-frequency signals in node representations and propose a graph convolutional network that can adaptively integrate different signals~\cite{2021Beyond}. Based on Simple Graph Convolution (SGC), a fast, scalable, and interpretable model that can adapt to homophilous and heterophilous graphs has been proposed by \citeauthor{2022Simplified}~\cite{2022Simplified}. \citeauthor{liu2023beyond} train the edge discriminator that judges whether the edge is homophily or heterophily through the structural and feature information and contrasts the dual-channel encodings obtained from the discriminated homophilous and heterophilous edges to learn the node representations~\cite{liu2023beyond}. \citeauthor{Kangzhao} construct high homophilous and high heterophilous graphs and used mixed filters based on the new graph to extract low-frequency and high-frequency information~\cite{Kangzhao}. 
However, all of these methods are single-view clustering methods, and they ignore the complementary information between different views.
In this work, we propose a hybrid graph filter with an adaptive mechanism related to the homophily degree, which can alleviate the dilemma of existing MVGC methods (Section~\ref{sec: Related MVGC}) that suffer from frustrated performance due to the loss of high-frequency information when confronted with heterophilous graphs.
Moreover, we obtain consensus embedding by weighted fusion of the node embeddings obtained from the adaptive hybrid graph filter for each view, which mines the complementary information between views facilitating the final clustering effect.

\section{Methodology}
\textbf{Notations}.
Let $\mathcal{G} = (\mathcal{V}, \mathcal{E})$ denote a graph, where $\mathcal{V}$ is the node set with $N = |\mathcal{V}|$, and $\mathcal{E} \subseteq \mathcal{V} \times \mathcal{V}$ is the edge set without self-loops. The feature matrix of the nodes is denoted as $\mathbf{X} \in \mathbb{R}^{N \times d}$.
% , where $x_i$ represents the $d$-dimensional feature vector of node $i$. 
The symmetric adjacency matrix of $\mathcal{G}$ is $\mathbf{A} \in \mathbb{R}^{N \times N}$, with elements $a_{ij} = 1$ if there exists an edge between node $i$ and node $j$, and $a_{ij} = 0$ otherwise. The objective of the multi-view graph clustering task is to categorize a set of $n$ nodes into $c$ distinct classes. In this study, we normalize the matrix $\mathbf{A}^v$ of each view as $\widetilde{\mathbf{A}}^v = (\mathbf{D}^v)^{-1} \mathbf{A}^v$, where the diagonal matrix $\mathbf{D}^v_{ii} = \sum_j a_{ij}^v$ represents the degree matrix. The normalized graph Laplacian matrix is defined as $\widetilde{\mathbf{L}}^v = \mathbf{I} - \widetilde{\mathbf{A}}^v$, where $\mathbf{I}$ denotes the identity matrix.

\begin{figure*}[t]
\centering
\includegraphics[width=0.88\textwidth]{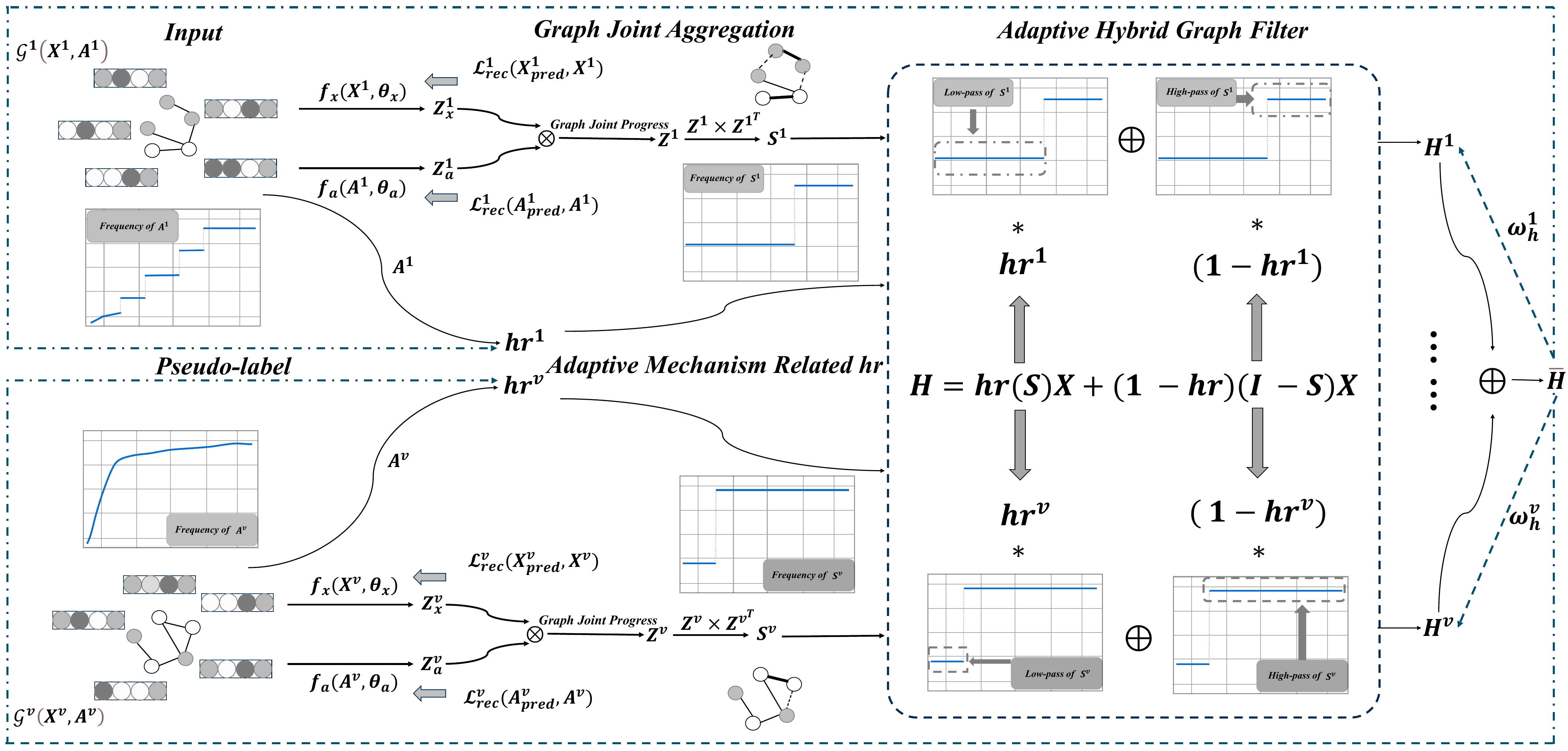}
\caption{The illustration of the proposed framework. The inputs to the framework are the feature matrix $\mathbf{X}$ and the adjacency matrix $\mathbf{A}$. The final output of the framework is the consensus embedding $\mathbf{\overline{H}}$.}
\label{fig:overview}
\end{figure*}

\subsection{Graph Joint Aggregation}\label{sec: Neighborhood Aggregation}
Suppose the input of each view includes both node features and adjacency relationships. In light of the inevitability of noise and the possible unreliability of adjacency relationships, we first explore $\mathbf{X}$ and $\mathbf{A}$ as follows:
\begin{equation}\label{eq: encoder}
\begin{gathered}
    \mathbf{Z}_x = f_x(\mathbf{X}; \theta_x), \quad 
    \mathbf{Z}_a = f_a(\mathbf{A}; \theta_a),
\end{gathered}
\end{equation}
where both $f_x(\cdot)$ and $f_a(\cdot)$ represent deep auto-encoder, $\theta_x$ and $\theta_a$ are learning parameters of the the respective autoencoders. $\mathbf{Z}_x \in \mathbb{R}^{N \times d'}$ and $\mathbf{Z}_a \in \mathbb{R}^{N \times d'}$ are the encoded outputs of $\mathbf{X}$ and $\mathbf{A}$ respectively, where $d' \leq d$ denotes the dimension after encoder dimensionality reduction. Empirically, $\mathbf{Z}_x$ and $\mathbf{Z}_a$ are more informative denoised hidden layer embeddings compared to the original feature and adjacency matrices, representing the extracted nodes' feature information and neighbor relationship information, respectively.

The input graph $\mathbf{A}$ may be homophilous, heterophilous, or mixed, and it is feasible to use it directly for filtering, but its dominance in the filtering operation may lead to excessive loss of node features information. Given this, we reconstruct a graph joint matrix by the following:
\begin{equation}\label{Neighborhood Aggregation}
    % \mathbf{Z} = \mathbf{Z}_a \times \mathbf{Z}_x^T .  \\
    \mathbf{Z} = \mathbf{Z}_a\mathbf{Z}_x^T .  \\
\end{equation}

Here, the obtained graph joint matrix $\mathbf{Z}$ is a similarity matrix that describes the similarity between $\mathbf{Z}_a$ and $\mathbf{Z}_x$. What's more, this graph joint matrix is made up of consensus from node features and neighborhoods, and it will be more trustworthy than using the initial input graph alone.

Furthermore, motivated by the message passing mechanism~\cite{scarselli2008graph}, where multi-order aggregation can smooth the graph signal, we aggregate the graph joint matrix $\mathbf{Z}$ to obtain the graph joint aggregation matrix $\mathbf{S}$ as follows:
\begin{equation}\label{eq: S}
    % \mathbf{S} = \mathbf{Z} \times \mathbf{Z}^T .
    \mathbf{S} = \mathbf{Z}\mathbf{Z}^T .
\end{equation}

Substitution of $\mathbf{Z}$ in Eq.~(\ref{eq: S}) by Eq.~(\ref{Neighborhood Aggregation}) yields: 
\begin{equation}\label{eq:aggregateS}
    % \mathbf{S} = (\mathbf{Z}_a \times \mathbf{Z}_x^T) \times (\mathbf{Z}_a \times\mathbf{Z}_x^T)^T , \\
    \mathbf{S} = (\mathbf{Z}_a\mathbf{Z}_x^T)(\mathbf{Z}_a\mathbf{Z}_x^T)^T , \\
\end{equation}
where the dimension of $\mathbf{S} \in \mathbb{R}^{N \times N}$ is the same as the original adjacency matrix $\mathbf{A}$, we regard graph joint aggregation matrix $\mathbf{S}$ as the modified graph. 

With Eq.~(\ref{eq:aggregateS}), the similarity in $\mathbf{Z}$ will be strengthened, and the low-frequency signal and the high-frequency signal will be effectively separated. 
%In other words, its frequency distribution will be sharpened. 
The results of ablation experiments and visualization of Fig.~\ref{fig:eigenvalues} reveal that the modified adjacency matrix $\mathbf{S}$ is more competitive than the original matrix $\mathbf{A}$. %Moreover, $\mathbf{S}$ serves the purpose of sharpening the frequencies.

\subsection{Homophily Degree Related Adaptive Hybrid Graph Filter}\label{sec: Adaptive Hybrid Graph Filters}

\subsubsection{Graph fourier transform.}The classic graph Laplacian, denoted as $\mathbf{L} = \mathbf{D} - \mathbf{A}$, is a symmetric positive semi-definite matrix~\cite{chung1997spectral}. Its eigendecomposition gives $\mathbf{L} = \mathbf{U}\mathbf{\Lambda} \mathbf{U}^T$, where $\mathbf{U}$ is the eigenvector matrix, also called \textit{graph Fourier basis}. Given a graph signal $x$, the graph Fourier transform of $x$ is defined as $\hat{x} = \mathbf{U}^Tx$, and the inverse graph Fourier transform is $x = \mathbf{U}\hat{x}$. The eigenvalue matrix $\mathbf{\Lambda} = \text{diag}(\lambda_1, \ldots, \lambda_N)$ with $\lambda_1 \leq \ldots \leq \lambda_N$ and $\lambda_i$ also called as \textit{frequency}~\cite{luan2020complete}. Hence, smaller $\lambda_i$ corresponds to low-frequency signals, whereas larger $\lambda_i$ corresponds to high-frequency signals. \citeauthor{dakovic2019local} also point out that $\lambda_i$ is smaller, implying that its corresponding basis function $u_i$ defined on the graph is smoother~\cite{dakovic2019local}. This indicates that low-frequency signals correspond to smooth information (similarity information) and high-frequency signals correspond to non-smooth information (dissimilarity information) on the graph.

\subsubsection{LP, HP and adaptive hybrid graph filter.} Empirically, low-pass and high-pass filters are associated with affinity and Laplacian matrices and their variants, respectively~\cite{luan2020complete}. In practice, there are two broad categories of the most widely used low-high-pass filters. One is built from symmetrically normalized affine matrix $\mathbf{A}_{sym}$ and Laplacian matrix $\mathbf{L}_{sym}$:
\begin{equation}
%\begin{aligned}
    \mathbf{LP} = \mathbf{A}_{\text{sym}} \mathbf{X}, \quad 
    \mathbf{HP} = \mathbf{L}_{\text{sym}} \mathbf{X},
%\end{aligned}
\end{equation}
where $\mathbf{A}_{sym} = \mathbf{D}^{-1/2}\mathbf{A}\mathbf{D}^{-1/2}$, $\mathbf{L}_{sym} = \mathbf{I} - \mathbf{A}_{sym}$, $\mathbf{LP}$ and $\mathbf{HP}$ represent the filtering results of the low-pass and high-pass filters, respectively.

The other is built from randomly wandering normalized affine matrix $\mathbf{A}_{rw}$ and the Laplacian matrix $\mathbf{L}_{rw}$:
\begin{equation}\label{eq:rw}
%\begin{aligned}
    \mathbf{LP} = \mathbf{A}_{\text{rw}} \mathbf{X}, \quad 
    \mathbf{HP} = \mathbf{L}_{\text{rw}} \mathbf{X},
%\end{aligned}
\end{equation}
where $\mathbf{A}_{rw} = \mathbf{D}^{-1}\mathbf{A}$, $\mathbf{L}_{rw} = \mathbf{I} - \mathbf{A}_{rw}$.

According to the graph Fourier transform introduced earlier, the convolution operation between the graph signal $x$ and the convolution kernel $f$ is as follows:
\begin{equation}
\begin{aligned}
    f * x &= \mathbf{U}((\mathbf{U}^Tf)\odot(\mathbf{U}^Tx)) = \mathbf{U}g^\theta \mathbf{U}^Tx,
\end{aligned}
\end{equation}
where $\odot$ denotes Hadamard product and $g^\theta = \mathbf{U}^Tf$ indicates the convolution kernel $f$ in the spectral domain.

If we use $\mathbf{A}_{rw}$ and $\mathbf{L}_{rw}$ in Eq.~(\ref{eq:rw}) as convolution kernels, respectively, then the graph signal $x$ will be filtered as:
\begin{equation}\label{eq: convolution kernels replace}
\begin{aligned}
    \mathbf{A}_{rw} * x &= \mathbf{U}(g_L^\theta)\mathbf{U}^Tx, \quad 
    \mathbf{L}_{rw} * x &= \mathbf{U}(g_H^\theta)\mathbf{U}^Tx,
\end{aligned}
\end{equation}
where $g_L^\theta$ and $g_H^\theta$ represent the $\mathbf{A}_{rw}$ and $\mathbf{L}_{rw}$ on the spectral domain respectively. $g_L^\theta = \mathbf{I} - \mathbf{\Lambda}$ and $g_H^\theta = \mathbf{\Lambda}$. Rewriting the results obtained from Eq.~(\ref{eq: convolution kernels replace}):
\begin{equation}\label{eq: projection}
\begin{aligned}
    \mathbf{U}(g_L^\theta)\mathbf{U}^Tx &= \sum_{i}(1 - \lambda_i)u_iu_i^Tx, \\
    \mathbf{U}(g_H^\theta)\mathbf{U}^Tx &= \sum_{i}\lambda_iu_iu_i^Tx.
\end{aligned}
\end{equation}

The effect of the filters is actually to adjust the scale of the components of $x$ in the frequency domain. Specifically, for a high-pass filter composed of Laplacian matrix with $\lambda_i \in [0, 2]$, larger eigenvalues ($\lambda_i > 1$) correspond to non-smooth eigenvectors $u_i$, smaller eigenvalues ($\lambda_i < 1$) are the opposite. The high-pass filter enhances the portion of the signal that fits well with non-smooth eigenvectors corresponding to large eigenvalues and depresses the signal that fits well with smooth eigenvectors corresponding to small eigenvalues. When $\lambda > 1$, the projection $\sum_{i}\lambda_iu_iu_i^Tx$ of Eq.~(\ref{eq: projection}) actually amplify the difference between $x_i$ and other non-smooth signals, when $\lambda < 1$, the projection $\sum_{i}\lambda_iu_iu_i^Tx$ further reduce differences between smooth signals~\cite{yang2023contrastive}, which demonstrates why the Laplacian matrix acts as a high-pass filter and shows that high-pass filter actually captures the dissimilarity information on the graph. On the other hand, a low-pass filter consisting of the affinity matrix with eigenvalues $(1 - \lambda_i) \in [-1, 1]$ only serves to smooth the signal due to eigenvalues $\lambda < 1$, $\textit{i.e.}$, it only captures the similarity information. 

Typically, the high-pass and low-pass filters are used to obtain the high-frequency and low-frequency information, respectively, on the graph. 
%From another perspective, the high-pass filter and low-pass filter actually learn the dissimilarities and similarities, respectively, in the graph. 
Importantly, the lack of either type of information leads to inadequate learning of node representations, which affects the downstream multi-view clustering task. In order to learn complete node representations without causing information loss, we design an adaptive hybrid graph filter as follows:
\begin{equation}\label{eq: adaptive hybrid filter}
\begin{gathered}
    \mathbf{H_{hybrid}} = hr \cdot (\mathbf{S}_{\text{rw}})^k\mathbf{X} + (1-hr) \cdot (\mathbf{I} - \mathbf{S}_{\text{rw}})^k\mathbf{X}, \\
    % \mathbf{LP} = (\mathbf{S}_{\text{rw}})^k\mathbf{X}, 
    % \mathbf{HP} = (\mathbf{I} - \mathbf{S}_{\text{rw}})^k\mathbf{X},
\end{gathered}
\end{equation}
where $\mathbf{H_{hybrid}}$ represents the output of the adaptive hybrid graph filter, $hr$ is a learnable parameter that measures the homophily degree and is used to control the adaptive process of the hybrid graph filter. $\mathbf{S}_{\text{rw}}$ is the graph joint aggregation matrix defined in Section~\ref{sec: Neighborhood Aggregation} with status equivalent to the affinity matrix normalized by random wandering $\mathbf{A_{rw}}$, $k$ is the order of the filters.

The low and high pass filters are combined with a tradeoff in the parameter $hr$ to form the adaptive hybrid graph filter. Low-pass and high-pass filters under the control of trade-off coefficient $hr$ adaptively acquire similarity and dissimilarity information on the graph to avoid information loss. 
\subsubsection{Adaptive mechanism related to homophily degree.}The homophily degree reflects the similarity of a given graph, furthermore, the homophily degree measures the proportion of similarity and dissimilarity information on the graph. The higher homophily degree indicates that the similarity information on the graph is dominant, and vice versa for the dissimilarity information. The adaptive mechanism of the adaptive hybrid graph filter we designed is motivated by the fact that the low-pass filter has to play a major role when similarity information is the dominant component on the graph, and the high-pass filter plays a major role when dissimilarity information is the dominant component. This fits perfectly with the intrinsic meaning of homophily degree. Naturally, we utilize the homophily degree to assign weights to the filters. However, the homophily degree of a given graph is difficult to measure due to the lack of true labels. Thus, we propose to learn the pseudo-label information~\cite{arazo2020pseudo} using the consensus embedding of the current iteration, and use the pseudo-label information and adjacency relationship information to compute the homophily ratio ($hr$):
\begin{equation}\label{hr}
    hr = \frac{\text{SUM}(\mathbf{A}^v \odot \mathbf{PP}^T - \mathbf{I})}{\text{SUM}(\mathbf{A}^v - \mathbf{I})},
\end{equation}
where, $\text{SUM}(\cdot)$ denotes the summation operation, $\odot$ denotes the Hadamard product, and $\mathbf{P} \in \{0, 1\}^{n\times c}$ is the one-hot encoding of the pseudo label. 

The $hr$ can also respond to the proportion of similarity and dissimilarity information on the graph, and it is a good substitute for the homophily degree due to its smaller calculation and complexity. 

Designing an adaptive mechanism using $hr$ has an additional advantage. Since $hr$ comes from the pseudo-labeling of the consensus embedding (Section~\ref{sec: Inter-view Embedding Fusion}) and the neighborhood information of the nodes, $hr$ actually becomes a mutually facilitating bridge between the consensus embedding and the adaptive hybrid graph filter. Specifically, when the adaptive hybrid graph filter under the control of $hr$ avoids information loss and learns distinguishable node embeddings, this in turn leads to better consensus embeddings obtained. The great consensus embedding will get more accurate pseudo-labeling information and homophily ratio, which will also facilitate the adaptive hybrid graph filter.

\subsection{View Weighting and Fusion}\label{sec: Inter-view Embedding Fusion}
In a multi-view task, different views contain not exactly the same information, $\textit{i.e.}$, there is consistency and complementarity among the views~\cite{jia2020semi, xu2022self, wang2022multi, cui2023novel, cui2023deep, ZhouZBZ23, zhou2024mcoco}. To fully utilize the complementary information between each view, we want to get a consensus embedding containing rich information by fusing the node embedding $\mathbf{H}^v$ of each view~\cite{Jia2021CoembeddingAS}. However, considering the different information values of different views, it is necessary for us to assign appropriate weights to each view after evaluating their qualities, which will control the different views to produce different contributions to the final consensus embedding. We first obtain the node embedding for each view by Eq.~(\ref{eq: adaptive hybrid filter}):
\begin{equation}\label{H^v}
    \mathbf{H}^v = hr^v \cdot (\mathbf{S}_{\text{rw}}^v)^k\mathbf{X}^v + (1-hr^v) \cdot (\mathbf{I} - \mathbf{S}_{\text{rw}}^v)^k\mathbf{X}^v .
\end{equation}
% thereby determining the fusion weights based on this evaluation. Ultimately, the weighted fused embedding is updated as follows:

Naturally, it occurs to us to utilize the obtained consensus embedding $\mathbf{\overline{H}}$ to in turn guide the embedding $\mathbf{H}^v$ of each view to assign weights to it. Specifically, if a view's embedding $\mathbf{H}^v$ is similar to the consensus embedding, then the information it carries must be important and we assign larger weight to it, and vice versa. We obtain the consensus embedding $\mathbf{\overline{H}}$ as follows:
\begin{equation} \label{eq:ovaH}
\begin{aligned}
    \mathbf{\overline{H}} = &\sum_{v=1}^V \omega_h^v \mathbf{H}^v, \text{where}\\ 
    & \omega_h^v = (\frac{eva^v}{\max{(eva^1, eva^2, \cdots, eva^V)}})^\rho,
\end{aligned}
\end{equation}
where $eva^v$ is obtained from the evaluation function that computes the similarity between the consensus embedding $\mathbf{\overline{H}}$ and each view embedding $\mathbf{H}^v$, $eva^v = evaluation(\mathbf{H}^v, \mathbf{\overline{H}})$. And the hyperparameter $\rho$ is used to adjust the degree of smoothing or sharpening of the view weights. For the final consensus embedding $\mathbf{\overline{H}}$, we apply the $k$-means algorithm to get the clustering results.

\subsection{Model Optimization}
In AHGFC, with the aim of gathering the maximum amount of precise information, we try to reconstruct $\mathbf{X}$ and $\mathbf{A}$ after obtaining $\mathbf{Z}_x$ and $\mathbf{Z}_a$ in Eq.~(\ref{eq: encoder}). The following is the reconstruction loss:
\begin{equation}\label{eq: rec}
    \mathcal{L}_{Rec} = l(f_x(\mathbf{X}; \theta_x); \mathbf{X}) + l(f_a(\mathbf{A}; \theta_a); \mathbf{A}),
\end{equation}
where $l(\cdot; \cdot)$ denotes loss function.

Learning from previous multi-view clustering studies~\cite{zhao2021graph, ren2022deep}, we also try to utilize Kullback-Leibler divergence loss ($\mathcal{L}_{KL}$) to improve AHGFC. We obtain the $\mathcal{L}_{KL}$ from the soft clustering distribution $\mathbf{Q}$ and the target distribution $\mathbf{P}$ as follows:
\begin{equation}\label{Lkl}
    \mathcal{L}_{KL} = \sum_{v=1}^V KL(\overline{\mathbf{P}} \Vert \mathbf{Q}^v) + \sum_{v=1}^V KL(\mathbf{P}^v \Vert \mathbf{Q}^v) + KL(\overline{\mathbf{P}} \Vert \overline{\mathbf{Q}}),
\end{equation}
where $q_{ij}^v \in \mathbf{Q}^v$ describes the probability that node $i$ in the $v$-th view belongs to the center of cluster $j$. $\mathbf{P}^v$ represents the target distribution of nodes embedding $\mathbf{H}^v$ in the $v$-th view. $\overline{\mathbf{Q}}$ and $\overline{\mathbf{P}}$ denote the soft and target distributions, respectively, of the consensus embedding $\mathbf{\overline{H}}$. $\mathcal{L}_{KL}$ encourages the soft distribution of each view to match the target distribution of the final consensus embedding $\mathbf{\overline{H}}$. Additionally, it enhances the consistency between the soft distribution and the target distribution of the consensus embedding.

Eventually, the loss of AHGFC is defined as:
\begin{equation}\label{eq: loss}
    \mathcal{L} = \gamma_{rec} \mathcal{L}_{Rec} + \gamma_{kl} \mathcal{L}_{KL},
\end{equation}
where $\gamma_{rec}$ and $\gamma_{kl}$ denote the trade-off coefficients. % for $\mathcal{L}_{Rec}$ and $\mathcal{L}_{KL}$, respectively.

\begin{table}
    \centering
    \setlength\tabcolsep{2.5pt}
    \begin{tabular}{cccccc}
    \toprule
        Datasets & Clusters & Nodes & Features & Graphs & $hr$ \\
        \midrule
         \multirow{2}*{Texas} & \multirow{2}*{$5$} & \multirow{2}*{$183$} & \multirow{2}*{$1703$} & $\mathcal{G}^1$ & $0.09$\\
         & & & & $\mathcal{G}^2$ & $0.09$ \\
        \midrule
         \multirow{2}*{Chameleon} & \multirow{2}*{$5$} & \multirow{2}*{$2277$} & \multirow{2}*{$2325$} & $\mathcal{G}^1$ & $0.23$\\
         & & & & $\mathcal{G}^2$ & $0.23$ \\
        \midrule
        \multirow{2}*{ACM} & \multirow{2}*{$3$} & \multirow{2}*{$3025$} & \multirow{2}*{$1830$} & $\mathcal{G}^1$ & $0.82$ \\
         & & & & $\mathcal{G}^2$ & $0.64$ \\
         \midrule
         {Wiki-cooc} & $5$ & $10000$ & $100$ & $\mathcal{G}^1$ & $0.34$ \\
         \midrule
         {Minesweeper} & $2$ & $10000$ & $7$ & $\mathcal{G}^1$ & $0.68$ \\
         \midrule
         {Workers} & $2$ & $11758$ & $10$ & $\mathcal{G}^1$ & $0.59$ \\
         \bottomrule
    \end{tabular}
    \caption{The statistics information of the six graph datasets.}\label{tab:datasets}
\end{table}

\begin{table*}[!ht]
\small
    \centering
    \resizebox{\linewidth}{!}{
    \begin{tabular}{r|cccc|cccc}
    \toprule[1.5pt]
    % \multirow{2}*{Methods / Datasets} & \multicolumn{4}{c|}{ACM (HR $0.82$ \& $0.64$)} & \multicolumn{4}{c}{Wiki-COOC (HR $0.34$)} \\
    \multirow{2}*{Methods / Datasets} & \multicolumn{4}{c|}{Texas ($hr$ $0.09$ \& $0.09$)} & \multicolumn{4}{c}{Chameleon ($hr$ $0.23$ \& $0.23$)} \\
         & NMI\% & ARI\% & ACC\% & F1\% & NMI\% & ARI\% & ACC\% & F1\% \\
    \midrule
    %Autoencoder & $19.4 \pm 0.7$ & $19.6  \pm 1.9$ & $52.7 \pm 2.3$ & $33.7 \pm 1.9$ & $21.6 \pm 0.04$ & $16.2 \pm 0.02$ & $44.02 \pm 0.04$ & $42.7 \pm 0.03$ \\
    VGAE (\citeyear{GAE}) & $12.7 \pm 4.4$ & $21.7 \pm 8.4$ & ${55.3} \pm 1.8$ & $29.5 \pm 3.1$ & $15.1 \pm 0.7$ & $12.4 \pm 0.6$ & $35.4 \pm 1.0$ & $29.6 \pm 1.7$ \\
    O2MAC (\citeyear{o2multi}) & $8.7 \pm 0.8$ & $14.6 \pm 1.8$ & $46.7 \pm 2.4$ & $29.1 \pm 2.4$ & $12.3 \pm 0.7$ & $8.9 \pm 1.2$ & $33.5 \pm 0.3$ & $28.6 \pm 0.2$ \\
    MvAGC (\citeyear{lin2021graph}) & $5.4 \pm 2.8$ & $1.1  \pm 4.1$ & $54.3 \pm 2.6$ & $19.8 \pm 5.1$ & $10.8 \pm 0.8$ & $3.3 \pm 1.7$ & $29.2 \pm 0.9$ & $24.3 \pm 0.5$ \\
    MCGC (\citeyear{pan2021multi}) & $12.7 \pm 2.9$ & $12.9 \pm 3.8$ & $51.9 \pm 0.9$ & $32.5 \pm 1.8$ & $9.5 \pm 1.3$ & $5.9 \pm 2.7$ & $30.0 \pm 2.0$ & $19.1 \pm 0.8$ \\
    % MVGC (\citeyear{XiaWYGHG22}) & $8.1 \pm 3.3$ & $7.8 \pm 3.1$ & $41.8 \pm 2.6$ & $28.4 \pm 3.1$ & $12.6 \pm 0.3$ & $5.1 \pm 0.6$ & $32.8 \pm 0.4$ & $26.9 \pm 0.5$ \\
    DuaLGR (\citeyear{ling2023dual}) & ${33.6 \pm 5.1}$ & ${24.1 \pm 3.7}$ & $55.4 \pm 2.1$ & ${42.0 \pm 2.0}$ & ${18.5 \pm 0.1}$ & ${13.6 \pm 0.1}$ & ${42.2 \pm 0.2}$ & ${41.1 \pm 0.2}$ \\
    AHGFC (ours) & $\mathbf{39.3 \pm 5.6}$ & $\mathbf{45.3 \pm 3.1}$ & $\mathbf{70.3 \pm 1.0}$ & $\mathbf{45.7 \pm 4.5}$ & $\mathbf{21.5 \pm 0.4}$ & $\mathbf{16.2 \pm 0.7}$ & $\mathbf{43.2 \pm 0.3}$ & $\mathbf{41.6 \pm 1.1}$ \\
    \midrule[1pt]
    Methods / Datasets & \multicolumn{4}{c|}{ACM ($hr$ $0.82$ \& $0.64$)} & \multicolumn{4}{c}{Wiki-cooc ($hr$ $0.34$)} \\
    \midrule
    % RMSC (\citeyear{xia14RMSC}) & $39.7$ & $33.1$ & $63.2$ & $57.5$ & $71.1$ & $76.5$ & $89.9$ & $82.5$ \\
    % LINE (\citeyear{LINE}) & $39.4$ & $34.3$ & $64.8$ & $65.9$ & $66.8$ & $69.9$ & $86.9$ & $85.5$ \\
    % VGAE (\citeyear{GAE}) & $49.1$ & $54.4$ & $82.2$ & $82.3$ & $69.3$ & $74.1$ & $88.6$ & $87.4$ \\
    % PMNE (\citeyear{PMNE}) & $46.5$ & $43.0$ & $69.4$ & $69.6$ & $59.1$ & $52.7$ & $79.3$ & $79.7$ \\
    % SwMC (\citeyear{nie2017SwMC}) & $8.4$ & $4.0$ & $41.6$ & $47.1$ & $37.6$ & $38.0$ & $65.4$ & $56.0$ \\
    % MNE (\citeyear{zhan2018MNE}) & $30.0$ & $24.9$ & $63.7$ & $64.8$ & $-$ & $-$ & $-$ & $-$ \\
    VGAE (\citeyear{GAE}) & $8.3 \pm 0.7$ & $5.0 \pm 0.4$ & $44.4 \pm 0.6$ & $43.8 \pm 0.7$ & $4.8 \pm 0.1$ & $12.8 \pm 0.2$ & $71.7 \pm 0.4$ & $\mathbf{61.3 \pm 0.1}$ \\
    O2MAC (\citeyear{o2multi}) & $67.2 \pm 1.1$ & $72.2 \pm 1.3$ & $89.7 \pm 0.5$ & $89.9 \pm 0.5$ & $38.0 \pm 5.3$ & $26.5 \pm 7.3$ & $58.6 \pm 3.5$ & $39.5 \pm 4.9$ \\
    MvAGC (\citeyear{lin2021graph}) & $57.8 \pm 1.3$ & $60.1 \pm 1.8$ & $84.4 \pm 0.9$ & $84.6 \pm 0.9$ & $30.8 \pm 6.8$ & $31.0 \pm 17.3$ & $62.5 \pm 7.2$ & $38.9 \pm 4.1$ \\
    MCGC (\citeyear{pan2021multi}) & $70.9 \pm 0.0$ & $76.6 \pm 0.0$ & $91.5 \pm 0.0$ & $91.6 \pm 0.0$ & ${15.7 \pm 3.3}$ & $7.4 \pm 6.2$ & ${47.2 \pm 6.7}$ & ${27.9 \pm 4.3}$ \\
    % MVGC (\citeyear{XiaWYGHG22}) & ${71.0 \pm 7.6}$ & ${75.5 \pm 10.7}$ & ${91.3 \pm 4.2}$ & ${91.2 \pm 4.3}$ & $--$ & $--$ & $--$ & $--$ \\
    %DCRN (\citeyear{DCRN}) & $\mathbf{72.2}$ & $\mathbf{78.1}$ & $92.1$ & $92.1$ & $49.4$ & $54.1$ & $79.9$ & $79.5$ \\
    % \midrule
    DuaLGR (\citeyear{ling2023dual}) & ${72.5 \pm 0.5}$ & ${78.8 \pm 0.5}$ & ${92.4 \pm 0.2}$ & ${92.5 \pm 0.2}$ & $5.4 \pm 0.7$ & $0.4 \pm 1.3$ & $32.2 \pm 1.2$ & $23.3 \pm 1.1$ \\
    AHGFC (ours) & $\mathbf{90.1 \pm 2.1}$ & $\mathbf{91.4 \pm 1.5}$ & $\mathbf{93.0 \pm 0.5}$ & $\mathbf{92.9 \pm 0.5}$ & $\mathbf{63.6 \pm 8.7}$ & $\mathbf{67.4 \pm 8.7}$ & $\mathbf{77.3 \pm 0.4}$ & ${60.6 \pm 5.5}$ \\
    \midrule[1pt]
    Methods / Datasets & \multicolumn{4}{c|}{Minesweeper ($hr$ $0.68$)} & \multicolumn{4}{c}{Workers ($hr$ $0.59$)} \\
    \midrule
    VGAE (\citeyear{GAE}) & $\mathbf{4.5 \pm 0.4}$ & $\mathbf{11.1 \pm 2.2}$ & $69.7 \pm 3.0$ & $\mathbf{60.1 \pm 1.2}$ & $\mathbf{3.4 \pm 0.3}$ & $\mathbf{11.7 \pm 0.5}$ & $73.7 \pm 0.9$ & $\mathbf{59.5 \pm 0.6}$ \\
    O2MAC (\citeyear{o2multi}) & $3.3 \pm 0.4$ & $2.8 \pm 1.2$ & $58.3 \pm 1.9$ & $53.9 \pm 1.3$ & $0.0 \pm 0.0$ & $0.1 \pm 0.1$ & $78.0 \pm 0.0$ & $44.3 \pm 0.0$ \\
    MvAGC (\citeyear{lin2021graph}) & $0.5 \pm 0.2$ & $-1.3  \pm 2.1$ & $58.8 \pm 3.1$ & $46.5 \pm 3.2$ & $4.7 \pm 0.2$ & $0.0 \pm 0.1$ & $54.7 \pm 0.1$ & $52.9 \pm 0.1$ \\
    MCGC (\citeyear{pan2021multi}) & $0.3 \pm 0.4 $& $-1.7 \pm 2.1$ & $66.3 \pm 7.7$ & $47.1 \pm 1.4$ & $0.0 \pm 0.1$ & $-0.8 \pm 0.7$ & $66.1 \pm 2.2$ & $49.1 \pm 0.8$ \\
    % MVGC (\citeyear{XiaWYGHG22}) & $-- \pm --$ & $-- \pm --$ & $-- \pm --$ & $-- \pm --$ & $-- \pm --$ & $-- \pm --$ & $-- \pm --$ & $-- \pm --$ \\
    DuaLGR (\citeyear{ling2023dual}) & ${0.2 \pm 0.2}$ & ${-0.3 \pm 1.1}$ & $60.0 \pm 1.9$ & ${47.8 \pm 1.9}$ & ${2.5 \pm 0.0}$ & ${0.9 \pm 0.1}$ & ${55.2 \pm 0.1}$ & ${52.2 \pm 0.1}$ \\
    AHGFC (ours) & ${0.0 \pm 0.1}$ & ${0.0 \pm 0.1}$ & $\mathbf{80.0 \pm 0.0}$ & ${44.8 \pm 0.6}$ & ${0.5 \pm 0.4}$ & ${0.9 \pm 0.9}$ & $\mathbf{78.3 \pm 0.1}$ & ${44.8 \pm 1.0} $ \\
    \bottomrule[1.5pt]
    \end{tabular}}
     \caption{The results of clustering on homophilous and heterophilous graph datasets. The best results are shown in bold.}
     \label{tab:overall_results}
\end{table*}

\section{Experiments}
\subsection{Experiments Setup}

\subsubsection{Datasets.}
We selected several representative datasets, \emph{i.e.}, homophilous graph dataset ACM and heterophilous graph datasets Texas and Chameleon~\cite{ling2023dual}. However, recent work~\cite{platonov2023critical} points out the shortcomings of low number of nodes and high number of duplicated nodes in Texas and Chameleon, respectively, and thus we additionally selected three new datasets: Wiki-cooc, Minesweeper and Workers~\cite{platonov2023critical}. 
Table~\ref{tab:datasets} summarizes the statistics of these six datasets. 
% and more details are given in Appendix~\ref{app:dataset}.
For all datasets, including homophilous graph datasets, heterophilous graph datasets, and new single-view graph datasets, we conducted five experiments and reported the mean and standard deviation of the results.

\subsubsection{Baselines.}
Several baselines are replicated for comparison with our model. VGAE~\cite{GAE} is a classical single-view clustering method. O2MAC~\cite{o2multi} is the method that learns from both node features and graphs. MvAGC~\cite{lin2021graph} and MCGC~\cite{pan2021multi} are two methods based on graph filters to learn a consensus graph for clustering. DualGR~\cite{ling2023dual} leverages soft-label and pseudo-label to guide the graph refinement and fusion process for clustering.

\subsubsection{Metrics.}
Following previous works, four commonly used metrics, \textit{i.e.}, normalized mutual information (NMI), adjusted rand index (ARI), accuracy (ACC), and F1-score (F1), are adopted to evaluate the clustering performance.
\subsubsection{Experimental environment.}
All experimental results were obtained in the same environment: NVIDIA GeForce GTX 1080Ti (GPU); 
CUDA version: 11.6; 
torch version: 1.13.1; 
12th Gen Intel(R) Core(TM) i7-12700KF (CPU).
\begin{table*}[!t]
    \centering
    \begin{tabular}{c|cccc|cccc}
    \toprule[1.5pt]
    \multirow{2}*{Compenents / Datasets} & \multicolumn{4}{c|}{Chameleon ($hr$ $0.23$ \& $0.23$)} & \multicolumn{4}{c}{Minesweeper ($hr$ $0.68$)} \\
         & NMI\% & ARI\% & ACC\% & F1\% & NMI\% & ARI\% & ACC\% & F1\% \\
    \midrule
    AHGFC (w/o $\mathcal{L}_{Rec}$) & $21.0$ & $16.2$ & $41.5$ & $37.2$ & $0.0$ & $-0.1$ & $63.7$ & $49.6$ \\
    AHGFC (w/o $\mathcal{L}_{KL}$)  & ${21.1}$ & $16.0$  & $43.2$ & ${41.6}$ & $0.0$ & $0.0$ & $80.0$ & $44.4$ \\
    \midrule
        AHGFC (w/o ${\mathbf{S}_a}$) & $14.1$  & $9.7$ & $35.0$ & $33.5$  & $0.0$  & ${0.2}$ & $69.4$ & $\mathbf{50.3}$ \\
         AHGFC (w/o $F_a$) & $6.5$  & $5.6$ & $34.1$ & $29.2$  & $0.0$ & $0.1$ & $69.8$ & ${50.1}$ \\
         AHGFC (w/o ${(\mathbf{S} \& F)_a}$) & $13.5$  & $8.8$ & $35.2$ & $34.3$  & $0.0$ & $-0.2$ & $63.2$ & ${49.3}$ \\
     \midrule
     \textbf{AHGFC} & $\mathbf{21.5}$ & $\mathbf{16.2}$ & $\mathbf{43.2}$ & $\mathbf{41.6}$ & $\mathbf{0.1}$ & $\mathbf{0.2}$ & $\mathbf{80.0}$ & ${44.8}$ \\
    \bottomrule[1.5pt]
    \end{tabular}
    \caption{The ablation study results of AHGFC on Chameleon and Minesweeper. The original results are shown in bold.
    $\mathbf{S}_a$ denotes the replacement of graph joint aggregation matrix $\mathbf{S}$ with adjacency matrix $\mathbf{A}$, $F_a$ denotes the replacement of adaptive hybrid graph filter with a common low-pass filter and $(\mathbf{S}\&F)_a$ denotes the combination of the above two.}
    \label{tab:ablation}
\end{table*}

\subsection{Overall Results}
Table~\ref{tab:overall_results} presents the results of all compared methods on homophilous and heterophilous graph datasets, from which we have the following observations. The experimental results on the homophilous graph dataset ACM demonstrate that AHGFC is highly competitive with the state-of-the-art models, with a slight increase in ACC, as well as significant improvements in NMI and ARI by 17.6\% and 12.6\% respectively. Moreover, AHGFC outperforms other baselines on heterophilous graph datasets, $i.e.$, Texas and Chameleon, which are typically challenging for most models. Specifically, when compared to SOTAs, AHGFC improves ACC by 14.9\% and ARI by 21.2\% on Texas. We also evaluate the performance of AHGFC on single-view datasets to address heterophilous issues by conducting experiments on three newly proposed single-view graph datasets~\cite{platonov2023critical}. Our model performs better than baselines in most of the metrics. Specifically, on Wiki-cooc dataset, AHGFC improves the best baseline ACC by 5.6\%, NMI by 25.6\%, and ARI by 36.4\%. Conclusively, AHGFC surpasses SOTAs on homophilous graphs and performs equally well on heterophilous graphs. This effectively mitigates the problem of multi-view graph clustering on heterophilous graphs.

\subsection{Ablation Studies and Analysis}
\subsubsection{Effect of each loss.}
To understand the importance of the reconstruction loss $\mathcal{L}_{Rec}$ of auto-encoder and the Kullback-Leibler
divergence loss $\mathcal{L}_{KL}$, we removed each loss respectively to observe the changes in performance. The experimental results are shown in Table~\ref{tab:ablation}. As can be seen, $\mathcal{L}_{Rec}$ dominates the total losses in terms of their impact on model performance, while the $\mathcal{L}_{KL}$ has almost no impact.

\subsubsection{Effect of each component.}
Graph joint aggregation matrix $\mathbf{S}$ and adaptive hybrid graph filter are important components of the AHGFC. We conducted three ablation experiments on both to analyze their impact on the performance of the AHGFC. 
% The related experimental results can be seen in Table~\ref{tab:ablation}. The experimental results show that the performance of the model is greatly and adversely affected whether the graph joint aggregation matrix or adaptive hybrid graph filter or both are eliminated. 
As Table~\ref{tab:ablation} demonstrates, the performance of the model is greatly and adversely affected whether the graph joint aggregation matrix or adaptive hybrid graph filter, or both are eliminated. 
Specifically, compared with the original model, the ACC of the ablation experiments on Chameleon decreases by at least 8\% and at most 9.1\%, while the ACC of the ablation experiments on Minesweeper decreases by at least 10.2\% and at most 16.8\%.

\begin{figure}[!h]
    \centering
    \subfigure{
        \includegraphics[height=1.233in,width=1.60in]{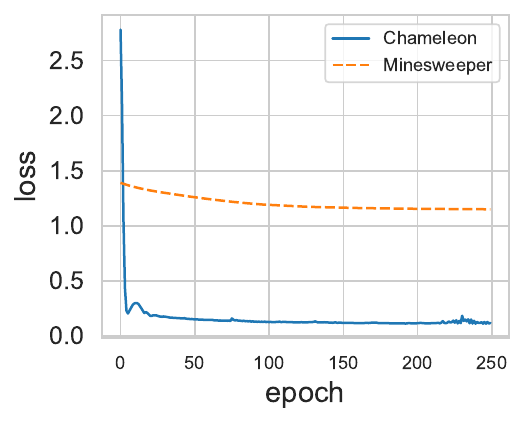}}
    %\hspace{0.5in}
    \subfigure{
        \includegraphics[height=1.233in,width=1.60in]{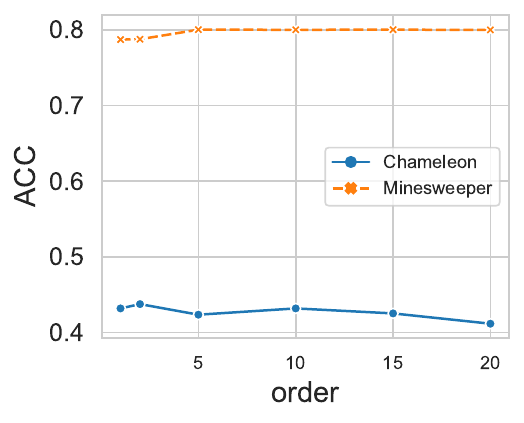}}
    \caption{Convergence analysis (left) and parameter sensitive analysis (right) on Chameleon and Minesweeper.}
    \label{fig:ablation}
\end{figure}

\subsubsection{Convergence analysis.}
The left part of Fig.~\ref{fig:ablation} shows the trend of loss on the Chameleon and Minesweeper. The initial value of loss on Chameleon is high but rapidly decreases to a very small value at the beginning. The value of loss on Minesweeper has a small but very smooth variation. Eventually, the losses in both datasets leveled off and converged.

% \begin{figure}[t]
%     \centering
%     \subfigure{
%         \includegraphics[height=1.2in,width=1.60in]{loss_epoch.pdf}}
%     %\hspace{0.5in}
%     \subfigure{
%         \includegraphics[height=1.2in,width=1.60in]{acc.pdf}}
%     \caption{Convergence analysis (left) and parameter sensitive analysis (right) on Chameleon and Minesweeper.}
%     \label{fig:ablation}
% \end{figure}

\subsubsection{Parameter sensitive analysis.}
The sensitivity analysis for $order$ is on the right of Fig.~\ref{fig:ablation}. From the spatial perspective, $order$ controls the aggregation order of graph filter. The higher $order$ enables nodes to aggregate information from more distant ones, while nodes can only access feature information of closer nodes in lower $order$. On the right of Fig.~\ref{fig:ablation}, Chameleon's ACC is larger at $order$ $=$ $\{1, 2, 10\}$, which is about 43.0\%, while it slightly decreases at $order$ $\geq$ $15$. We speculate that the abundance of duplicated nodes in Chameleon may limit the improvement in node representation with higher-order neighboring nodes' feature information. However, overall, Chameleon's ACC does not change much between different orders, which reflects that AHGFC can learn distinguishable node representations at different orders, while Minesweeper's ACC stays at 80.0\% at $order$ $\geq$ $5$, which also corroborates this statement.

\section{Conclusion}
% In this work, we analyze the reasons why MVGC methods based on traditional GNNs are frustrated when dealing with heterophilous graphs from graph filtering perspectives and thus find a breakthrough by proposing an adaptive hybrid graph filter for multi-view graph clustering (AHGFC). 
In this study, we analyze challenges faced by traditional GNNs-based MVGC methods in handling heterophilous graphs from a graph filtering perspective. Our breakthrough solution is the introduction of an adaptive hybrid graph filter for multi-view graph clustering (AHGFC).
Specifically, we design an adaptive hybrid graph filter related to homophily degree, which can adaptively mine low and high-frequency information in the graph to learn distinguishable node embeddings. 
% In addition, we construct the filter by applying a graph joint aggregation matrix that can enhance distinguishability between low-frequency and high-frequency signals instead of directly employing initial adjacency matrix. 
Additionally, we build the filter using a graph joint aggregation matrix to enhance the distinguishability between low and high-frequency signals, instead of directly using initial adjacency matrix.
We evaluate AHGFC's performance on both multi-view homophilous and heterophilous graph datasets, as well as on single-view graph datasets. 
The results demonstrate AHGFC's effective mitigation of heterophilous graph challenges in MVGC, while maintaining competitive performance on homophilous graphs.

\section{Acknowledgments}
This work is supported in part by National Key Research and Development Program of China (Nos. 2020YFC2004300 and 2020YFC2004302), National Natural Science Foundation of China (No. 61971052), and Shenzhen Science and Technology Program (Nos. JCYJ20230807120010021 and JCYJ20230807115959041). Lifang He is partially supported by the NSF grants (MRI-2215789, IIS-1909879, IIS-2319451), NIH grant under R21EY034179, and Lehigh's grants under Accelerator and CORE.
\bibliography{aaai24}

\end{document}